\documentclass[runningheads]{llncs}
\usepackage[T1]{fontenc}
\usepackage{graphicx}
\usepackage{hyperref}       % hyperlinks
\usepackage{url}            % simple URL typesetting
\usepackage{booktabs}       % professional-quality tables
\usepackage{nicefrac}       % compact symbols for 1/2, etc.
\usepackage{microtype}      % microtypography
\usepackage{lipsum}
\usepackage{multirow}
\usepackage{comment}

\usepackage{amsfonts,amssymb,bm}
\usepackage{mathtools}
\usepackage{textcomp}
\usepackage{multicol}
\usepackage{bbm}
\usepackage{hyperref}
\usepackage{placeins}
\usepackage{chngpage}
\usepackage{caption}
\usepackage{subcaption}
\usepackage{multicol}
\usepackage{multirow}
\usepackage{booktabs}
\usepackage{adjustbox}
\usepackage{algorithmicx}
\usepackage{algorithm}
\usepackage{algpseudocode}
\algrenewcommand\algorithmicindent{1.0em}

\usepackage{amsmath}
\usepackage{float}

\usepackage{color}

\begin{document}
\title{Robot Navigation with Entity-Based Collision Avoidance using Deep Reinforcement Learning}
\titlerunning{Entity-Based Collision Avoidance using Deep Reinforcement Learning}

\author{
Yury Kolomeytsev \and 
Dmitry Golembiovsky
}
\authorrunning{Y. Kolomeytsev \and D. Golembiovsky}
\institute{
Lomonosov Moscow State University, Moscow, Russia \\
\email{yury.kolomeytsev@gmail.com}
}
\maketitle              % typeset the header of the contribution

\begin{abstract}
Efficient navigation in dynamic environments is crucial for autonomous robots interacting with moving agents and static obstacles. We present a novel deep reinforcement learning approach that improves robot navigation and interaction with different types of agents and obstacles based on specific safety requirements. Our approach uses information about the entity types, improving collision avoidance and ensuring safer navigation. We introduce a new reward function that penalizes the robot for being close to or colliding with different entities such as adults, bicyclists, children, and static obstacles, while also encouraging the robot's progress toward the goal. We propose an optimized algorithm that significantly accelerates the training, validation, and testing phases, enabling efficient learning in complex environments. Comprehensive experiments demonstrate that our approach consistently outperforms state-of-the-art navigation and collision avoidance methods.

\end{abstract}

\section{Introduction}
The development of robots, designed to assist humans, has revolutionized numerous sectors, including industrial automation, manufacturing, agriculture, space, medical science, household utilities, delivery, and social services \cite{AutonomousRobots}.
Autonomous navigation, a key attribute of these robots, enables them to move independently, perceive the environment, and take appropriate actions to complete tasks successfully, safely and efficiently in terms of computational and energy resources.

Recent research interest in mobile robots, particularly in the context of applications such as delivery, search and rescue, service, and warehouse robots has emphasized the importance of robot navigation \cite{drew2021multi,KRUSE20131726}.
The core challenge is planning a path to a destination while avoiding dynamic agents and static obstacles \cite{lavalle_2006}, a task that is especially difficult in crowded areas. Traditional navigation methods often treat moving agents as static obstacles or use short-sighted reactions, leading to unsafe and unnatural robot behavior \cite{SocialAwareRobot_2013,UnfreezingtheRobot_2010}.

Navigating in a socially compliant manner in populated, dynamic environments is a significant challenge \cite{CHARALAMPOUS201785}. A robot must account for the diversity of entities around it, as each has unique characteristics, movements and safety requirements. We use the term ``environmental entities'', or simply entities, to describe both moving agents and static obstacles. For instance, a sidewalk autonomous delivery robot \cite{GEHRKE2023100789} interacts with various moving agents (e.g., adults, children, bicyclists) and static obstacles (e.g., curbs, trees).

For robots to navigate in densely populated spaces in a socially compliant manner, it is crucial to understand the behavior of agents and obstacles and adhere to cooperative rules. To achieve that, we implement a new method, based on deep reinforcement learning (RL), which allows the robot to navigate in dynamic environments and interact differently with different types of agents and obstacles based on various safety requirements. We call this approach \emph{Entity-Based Collision Avoidance using Deep Reinforcement Learning} (EB-CADRL).

Our main contributions are as follows. We propose a new method of collision avoidance based on deep RL which can effectively utilize information about the types of entities, which helps the robot to avoid collisions and leads to safer interaction and navigation. We propose a reward function that penalizes the robot differently depending on the type of agent or obstacle it collides with and encourages the robot to get close to the goal. Moreover, our new reward function penalizes the robot for being close to entities and the safe distance also depends on the entity. We introduce an optimized algorithm for training and testing the model which significantly accelerates the speed of training and testing and allows the robot to be trained effectively in complex environments. We test the applicability and effectiveness of the proposed approach in our simulator. Our experiments demonstrate that the new approach consistently outperforms standard methods of robot navigation with collision avoidance, including state-of-the-art methods.

\section{Related Work}\label{sec:relwork}

Extensive research has led to significant advancements in robot navigation including navigation in crowded environments.

Traditional methods include the Social Force Model (SFM) \cite{social_force}, which uses attractive and repulsive forces, and Optimal Reciprocal Collision Avoidance (ORCA) \cite{ORCA}, which uses velocity obstacles to find collision-free paths.
These approaches rely on hand-crafted rules and often fail when their assumptions are not met in real social situations.

Recently, learning-based methods have gained significant attention. Several studies use imitation learning to derive navigation strategies from demonstrations of desired behaviors.
In \cite{navigation_raw_depth_inputs_2018} the authors use generative adversarial imitation learning to develop a navigation policy that operates directly on raw depth images.
In \cite{Kretzschmar_2016} maximum entropy inverse RL has been applied  to extract underlying cooperation features from demonstrated pedestrian trajectories. However, the effectiveness of these methods heavily relies on the scale and quality of the demonstrations, which can be resource-intensive and limit the policy's quality.

Deep reinforcement learning (DRL) is another prominent approach, treating social navigation as a Markov decision process \cite{socially_aware_2017}. Early DRL methods like CADRL \cite{CADRL} used value networks based on agents' joint states, later enhanced by LSTM-RL \cite{LSTM-RL} to handle a variable number of agents. These methods, however, exhibited limitations in modeling complex crowd interactions. SARL \cite{SARL} improved on this by modeling both human-robot and human-human interactions, but it focused on spatial relationships while neglecting temporal dynamics. More recently, \cite{Xue2024} proposed a DRL framework using spatial-temporal reasoning to better understand crowd behavior and ensure safety. However, none of the aforementioned approaches addresses the challenge of navigating in environments populated by different types of entities.

In our work, we aim to address the limitations of the methods mentioned above by creating a more robust, safe, and socially compliant navigation method using deep reinforcement learning.

\section{Problem Formulation}\label{sec:problem_formulation}

We address the problem of autonomous robot navigation towards a goal in dynamic environments populated with various entities, including dynamic agents and static obstacles. This problem can be formulated as a Markov decision process (MDP).
The robot must navigate in a socially compliant manner, avoiding collisions while reaching its goal. We define the state of each agent (robot or environmental entity) in terms of observable and unobservable components:

\begin{itemize}
\item Observable states: position $\mathbf{p} = [p_x, p_y]$, velocity $\mathbf{v} = [v_x, v_y]$, and radius $r$.
\item Unobservable states: goal position $\mathbf{g} = [g_x, g_y]$, preferred speed $v_{\text{\text{pref}}}$, and heading angle $\theta$.
\end{itemize}

While acting in the environment, the robot knows its observable and unobservable states, as well as the observable states of all entities. We use the robot-centric parameterization described in \cite{CADRL} and \cite{LSTM-RL}, where the robot is positioned at the origin and the x-axis is aligned with the robot's goal direction. Denote $e_i$ as a type of entity $i$. The robot’s state and the $i$-th entity's observable state at time $t$ could be represented as:
\begin{equation}
\begin{aligned}
    \mathbf{s}_t^r &= [d_g, v_x, v_y, r, v_{\text{\text{pref}}}, \theta], \\
    \mathbf{s}_t^{i o} &= [p_x^i, p_y^i, v_x^i, v_y^i, r^{i}, d^i, r^{i} + r, e_i],
\end{aligned}
\label{eq:states}
\end{equation}
where \(d_g = \|\mathbf{p}_t - \mathbf{g}\|_2\) is the distance from the robot to the goal, and \(d^i = \|\mathbf{p}_t - \mathbf{p}_t^i\|_2\) is the distance from the robot to the $i$-th entity.

The robot can acquire its own state and the observable states of other agents at each time step. The joint state at time $t$ is defined by:
\begin{equation}
\mathbf{s}_t^j = [\mathbf{s}_t^r, \mathbf{s}_t^{1 o}, \mathbf{s}_t^{2 o}, \ldots, \mathbf{s}_t^{n o}].
\label{eq:states_2}
\end{equation}

The robot’s velocity \(\mathbf{v}_t\) is determined by the action command \(\mathbf{a}_t\) from the navigation policy, i.e., \(\mathbf{v}_t = \mathbf{a}_t = \pi(\mathbf{s}_t^j)\).

The reward function \(R(\mathbf{s}_t^j, \mathbf{a}_t)\) is defined as a mapping from the current state \(\mathbf{s}_t^j\) and action \(\mathbf{a}_t\) to a scalar value, which represents the immediate benefit of taking action \(\mathbf{a}_t\) in state \(\mathbf{s}_t^j\). Formally:
\begin{equation}
R(\mathbf{s}_t^j, \mathbf{a}_t) = r_t,
\label{eq:reward}
\end{equation}
where \(r_t\) is the reward received at time step \(t\).

Our objective is to find the optimal policy \(\pi^*(\mathbf{s}_t^j)\) that maximizes the expected reward.

The optimal value function is given by \cite{Sutton}:
\begin{equation}
V^*(\mathbf{s}_t^j) = \sum_{\tilde{t}=t}^T \gamma^{\tilde{t} \cdot v_{\text{\text{pref}}}} R_{\tilde{t}}(\mathbf{s}_{\tilde{t}}^j, \pi^*(\mathbf{s}_{\tilde{t}}^j)),
\label{eq:optimal_value_func}
\end{equation}
where \(\gamma \in (0, 1)\) is a discount factor and $T$ is the time step at which the episode ends.

Using value iteration, the optimal policy is derived as:

\begin{equation}
\begin{aligned}
& \pi^*\left(\mathbf{s}_t^j\right)=\arg \max _{\mathbf{a}_t \in \mathbf{A}} \biggl[ R\left(\mathbf{s}_t^j, \mathbf{a}_t\right)+\gamma^{\Delta t \cdot v_{\text {pref }}} \cdot \\
& \quad \int_{\mathbf{s}_{t+\Delta t}^j} P\left(\mathbf{s}_{t+\Delta t}^j \mid \mathbf{s}_t^j, \mathbf{a}_t\right) V^*\left(\mathbf{s}_{t+\Delta t}^j\right) d \mathbf{s}_{t+\Delta t}^j \biggr],
\end{aligned}
\label{eq:optimal_policy}
\end{equation}

where $\Delta t$ is time step duration, \(R(\mathbf{s}_t^j, \mathbf{a}_t)\) is the reward function, \(\mathbf{A}\) is the action space, and \(P(\mathbf{s}_{t+\Delta t}^j \mid \mathbf{s}_t^j, \mathbf{a}_t)\) is the transition probability.

\section{Proposed Approach}\label{sec:new_approach}

In this section, we formulate our new approach which we call Entity-Based Collision Avoidance using Deep Reinforcement Learning \mbox{(EB-CADRL)}.

\subsection{Reward Function}\label{subsec:reward_function}
Assume the robot operates in an environment with entities of $N$ different types. Let $\mathcal{E} = \{e_1, e_2, \ldots, e_N\}$ represent the set of entity types. Define $d(e_i, t)$ as the minimum distance between the robot and any entity of type $e_i \in \mathcal{E}$ over the period $[t - \Delta t, t]$, where $i = 1, \ldots, N$.

We define $d_{\text{min}}(t) = \min_{i\in\{1,\dots,N\}} d(e_i, t)$ as the minimum distance between the robot and all entities around it over the period $[t - \Delta t, t]$. If a collision occurs with an entity of type $e_i$, then $d(e_i, t)$ and $d_{\text{min}}(t)$ are less than or equal to zero.

The function $R_{\text{time}}(t)$ represents the time reward:
\begin{equation}
R_{\text{time}}(t) = 
\begin{cases}
1 & \text{if } t < t_{\text{pref}} \\
\frac{t_{\text{max}} - t}{t_{\text{max}} - t_{\text{pref}}} & \text{if } t_{\text{pref}} \leq t \leq t_{\text{max}} \\
0 & \text{otherwise}
\end{cases}
\label{eq:time_reward}
\end{equation}
where $t_{\text{pref}}$ is the preferred time duration for reaching the goal, $t_{\text{max}}$ is the maximum allowable time to reach the goal.

Let $d_{\text{max}}$ be the maximum distance to the goal, which is the Euclidean distance between the starting position of the robot at time $t = 0$ and the position of the goal. Let $d_g(t)$ be the distance to the goal at time $t$. Then proximity reward can be represented as:
\begin{equation}
\begin{aligned}
R_{\text{prox}}(t) = 1 - \frac{d_g (t)}{d_{\text{max}}}.
\end{aligned}
\label{eq:proximity_reward}
\end{equation}

Define $R_{\text{coll}}(e_i)$ as the penalty associated with collisions involving an entity of type $e_i$.

Let $d_{\text{disc}}(e_i)$ be the discomfort distance for entity of type $e_i$ and $p_{\text{disc}}(e_i)$ be the discomfort penalty factor for entity of type $e_i$, $i = 1, \ldots, N$. We can then construct the penalty for being close to the entity:
\begin{equation}
c_{\text{disc}}(e_i, t) = (d(e_i, t) - d_{\text{disc}}(e_i)) * p_{\text{disc}}(e_i) * \Delta t.
\label{eq:close_to_entity}
\end{equation}

Let $i^{*}(t) \;=\;\arg\min_{i\in\{1,\dots,N\}} d(e_i, t)$.
For brevity, we use \(i^*\) for \(i^{*}(t)\). 
Then $e_{i^*} \in \mathcal{E}$ represents the entity type of the entity that was closest to the robot over the period $[t - \Delta t, t]$, meaning $d_{\text{min}}(t) = d(e_{i^*}, t)$.
Now, our new reward function can be represented as:

\begin{equation}
\begin{aligned}
R(t) =
\begin{cases}
R_{\text{prox}}(t) & \text {if } t \geq t_{\text{max}} \text { and } \mathbf{p}_{t} \neq  \mathbf{g}\\
R_{\text{coll}}(e_{i^*}) + R_{\text{prox}}(t) & \text {else if } d_{\text{min}}(t) \leq 0 \\
1 + R_{\text{time}}(t) & \text {else if } \mathbf{p}_{t}=\mathbf{g} \\
c_{\text{disc}}(e_{i^*}, t) & \text {else if } d(e_{i^*}, t) < d_{\text{disc}}(e_{i^*}) \\ 
0 & \text {otherwise }
\end{cases}
\end{aligned}
\label{eq:reward_function}
\end{equation}

The assignment of collision penalties, discomfort penalties, and discomfort distance thresholds in our model is based on the potential risk and harm associated with the robot's interactions with different entities.
% For instance, a collision with a static obstacle like a fence, wall or trash can might result in minimal damage to the robot with no risk to other entities. On the other hand, collisions with children could cause serious injury to them. Collisions with bicyclists could lead to serious injury for both the rider and the robot and could cause the rider to fall from the bicycle, increasing the potential harm. Our model, therefore, seeks to minimize these risks by training it to avoid collisions and dangerous situations.

\subsection{Model}\label{subsec:model}

Our model builds upon the architecture described in \cite{SARL}, enhancing it for the specific task of controlling a robot navigating in dynamic environments with various obstacles and pedestrians. To account for different types of entities (e.g., adults, children, bicycles), we incorporated an additional module. Initially, this module uses one-hot encoding to represent each agent type. For more advanced scenarios, one could use trainable embeddings for each agent type. These encoded features or embeddings are then concatenated with the primary features of the agents and passed into the model.

We embed the state of each entity \(i\), along with the state of the robot, into a fixed-length vector \(g_i\) using a multi-layer perceptron (MLP):
\begin{equation}
\begin{aligned}
g_i = \phi_g(s^r, s^{io}; W_g),
\end{aligned}
\label{eq:mlp}
\end{equation}
where \(\phi_g(\cdot)\) is an embedding function with rectified linear unit (ReLU) activations, and \(W_g\) are the embedding weights.

The embedding vector \(g_i\) is then fed into another MLP to obtain the pairwise interaction feature between the robot and entity \(i\):
\begin{equation}
\begin{aligned}
h_i = \psi_h(g_i; W_h),
\end{aligned}
\label{eq:pairwise_interaction}
\end{equation}
where \(\psi_h(\cdot)\) is a fully-connected layer with ReLU nonlinearity, and \(W_h\) are the network weights.

To handle varying numbers of entities, we use a social attentive pooling module \cite{SARL} which learns the relative importance of each neighbor and the collective impact of the crowd in a data-driven manner. The interaction embedding \(g_i\) is transformed into an attention score \(\alpha_i\) as follows:
\begin{equation} g_m = \frac{1}{n} \sum_{k=1}^n g_k, \qquad \alpha_i = \psi_\alpha(g_i, g_m; W_\alpha), \label{eq:attention_score2} \end{equation}
where \(g_m\) is a fixed-length embedding vector obtained by mean pooling all the individuals, \(\psi_\alpha(\cdot)\) is an MLP with ReLU activations, and \(W_\alpha\) are the weights.

Given the pairwise interaction vector \(h_i\) and the corresponding attention score \(\alpha_i\) for each neighbor \(i\), the final representation of the crowd is a weighted linear combination of all pairs:
\begin{equation}
\begin{aligned}
c = \sum_{i=1}^n \operatorname{softmax}(\alpha_i) h_i,
\end{aligned}
\label{eq:crowd_representation}
\end{equation}
where $\operatorname{softmax}(\alpha_i) = \frac{e^{\alpha_i}}{\sum_{j=1}^n e^{\alpha_j}}.$

Using the compact representation of the crowd $c$, we construct a planning module that estimates the state value $v$ for cooperative planning:
\begin{equation}
\begin{aligned}
v = f_v(s^r, c; W_v),
\end{aligned}
\label{eq:state_value_v_for_cooperative_planning}
\end{equation}
where \(f_v(\cdot)\) is an MLP with ReLU activations, and \(W_v\) are the weights. We denote this value network as \(V\).

\subsection{Algorithm Parallel Deep V-learning}\label{subsec:algorithm_v_learning}

The value network is trained by the temporal-difference method with standard experience replay and fixed target network techniques \cite{mnih2015humanlevel}. However, we improve it by using parallelism. We run several agents in several environments in parallel which helps us to collect the experience much faster than in the previous work. We call this approach ``Parallel Deep V-learning'', the pseudocode is shown in Algorithm \ref{algorithm_1} and Algorithm \ref{algorithm_2}.

On a 24-core CPU, our proposed Parallel Deep V-learning algorithm was 12 times faster than conventional Deep V-learning \cite{SARL,CADRL} in both training and evaluation. This significant speed-up allows for rapidly training more sophisticated models in environments with numerous dynamic agents and static obstacles.

\begin{minipage}{0.97\linewidth}

\begin{algorithm}[H]
\caption{Parallel Deep V-learning}\label{algorithm_1}
\begin{algorithmic}[1]
\State Run imitation learning with demonstration $D$, update value network $V$ with $D$
\State Initialize: target value network $\hat{V} \leftarrow V$, experience replay memory $E \leftarrow D$, parallel processes number $N$
\While{episode $<$ M}
    
    \State Initialize a multiprocessing Pool of size $N$
    \For{$k$ = episode, episode + $N$} in parallel:
    \State Run the function $RUN\_EPISODE(k)$
    \EndFor
    \State Sample random minibatch tuples from $E$
    \State Update value network $V$ by gradient descent
    \If{episode $\mod$ UpdateInterval = 0}
        \State Update target value network $\widetilde{V} \leftarrow V$
    \EndIf
    \State Update $episode \leftarrow episode + N$
\EndWhile
\State \Return $V$
\end{algorithmic}
\end{algorithm}

\end{minipage}

\vspace{-1.6em}

\begin{minipage}{0.97\linewidth}
\begin{algorithm}[H]
\caption{RUN\_EPISODE function}\label{algorithm_2}
\begin{algorithmic}[1]
\Function{RUN\_EPISODE}{k}
\State Initialize: environment, robot with current model $V$, random sequence $s_0^{j}$
\State Initialize temporary buffer $S$
\Repeat
    \State $\mathbf{a}_t \leftarrow \begin{cases} \text{RandomAction}(), \qquad \qquad \qquad \qquad \qquad \qquad \epsilon \\ \arg \max _{\mathbf{a}_t \in \mathbf{A}} R(\mathbf{s}_t^j, \mathbf{a}_t) + \gamma^{\Delta t \cdot v_{\text{\text{pref}}}} V(\mathbf{s}_{t+\Delta t}^{j}), \quad \space 1-\epsilon \end{cases}$
    \State Make step using $\mathbf{a}_t$
    \State Store tuple $(s_t^{j}, a_t, r_t, s_{t+\Delta t}^{j})$ in $S$
    \State Update time $t \leftarrow t + \Delta t$
\Until{$s_t$ is a terminal state or $t \geq t_{\max}$}
\If{Need to update memory}
    \For{$t = 0$, length($S$)}
    \State Get $r_t$, $s_t^{j}$ and $s_{t+1}^{j}$ from $S$
    \State Set $value_t = r_t + \gamma^{\Delta t \cdot v_{\text{\text{pref}}}} \hat{V}(s_{t+1}^{j})$
    \State Store ($s_t$, $value_t$) in $E$
    \EndFor
\EndIf
\EndFunction
\end{algorithmic}
\end{algorithm}

\end{minipage}

\subsection{Implementation Details}\label{subsec:impl_details}

We assume holonomic kinematics for the robot, i.e., it can move in any direction.
The action space consists of 81 discrete actions: a stop action (0 velocity) plus combinations of 5 speeds exponentially spaced between (0, $v_{\text{pref}}$] and 16 headings evenly spaced between $[0, 2\pi)$.

The neural network architecture consists of four multi-layer perceptrons (MLP). An MLP's structure is denoted by the tuple $(d_{\text{in}}, d_1, \dots, d_{\text{out}})$, which specifies the number of neurons in the input, hidden, and output layers, respectively. The dimensions of these MLP components are $\phi_g$: (17, 300, 200), $\psi_h$: (200, 200, 100), $\psi_{\alpha}$: (400, 200, 200, 1), and $f_v$: (106, 300, 200, 200, 1). 
The input dimension 17 to $\phi_g$ represents the joint state of the robot's state (6), agent's state (7), and entity type information (4), while the 106 input dimensions to $f_v$ combine the robot's state (6) with the attention-weighted feature representation of all agents (100).

We implemented the policy in the deep learning framework PyTorch \cite{Pytorch} and trained it with a batch size of 100 (for each episode) using stochastic gradient descent. For imitation learning, we collected 3000 episodes of demonstration using ORCA and trained the policy 50 epochs with a learning rate of 0.01. For reinforcement learning, the learning rate is 0.001 and the discount factor $\gamma$ is 0.9. The exploration rate of the $\epsilon$-greedy policy decays linearly from 0.5 to 0.05 in the first 25000 episodes and stays at 0.05 for the remaining episodes.

All input features, including the robot's state, entity states, and entity types, would be provided by localization and perception systems on a real robot. For our experiments, this information is obtained directly from the simulator's state, while the implementation of these systems on a physical robot is assumed to rely on existing algorithms \cite{SpringerHandbook} and is outside the scope of this research.

\section{Experiments}\label{sec:experiments}

\subsection{Reward setup}\label{reward_setup}
In this subsection we specify the reward function setup for our experiments, where the robot navigates in the environment containing 4 different types of entities: adults, bicycles, children, and static obstacles. The number of each entity type varies, with experiments involving between 3 and 9 of each, creating crowded scenarios with over 26 agents and obstacles.

We denote the entity types as follows: Adult ($A$), Bicycle ($B$), Child ($C$), and Obstacle ($O$). The penalties for collisions are defined as:

\begin{equation}
\begin{aligned}
R_{\text{coll}}(e) = 
\begin{cases}
-0.5, & \text{if } e = O \\
-1.0, & \text{if } e = A \\
-1.5, & \text{if } e = B \\
-2.0, & \text{if } e = C 
\end{cases}
\end{aligned}
\label{eq:penalties_for_collisions}
\end{equation}

The rationale behind the collision penalties is based on the predictability of the entity's behavior and the severity of the consequences of a collision. Static obstacles are the most predictable and have the least severe consequences, resulting in the smallest penalty (-0.5). Adults exhibit predictable behavior with moderate collision consequences, resulting in a penalty (-1.0). Cyclists, due to their higher speed and the potential for severe consequences such as falling off a bicycle, incur a higher penalty (-1.5). Children are the most unpredictable and vulnerable, leading to the most severe consequences in the event of a collision resulting in the highest penalty (-2.0).

Let
$d_{\text{disc}}(e)$ be the discomfort distance (in meters) for entity $e$:
\begin{equation}
\begin{aligned}
d_{\text{disc}}(e) = 
\begin{cases}
0.1, & \text{if } e = A \\
0.2, & \text{if } e = B \\
0.2, & \text{if } e = C
\end{cases}
\end{aligned}
\label{eq:discomfort_distance}
\end{equation}

Let
$p_{\text{disc}}(e)$ be the discomfort penalty factor for entity $e$:
\begin{equation}
\begin{aligned}
p_{\text{disc}}(e) = 
\begin{cases}
0.5, & \text{if } e = A \\
1.0, & \text{if } e = B \\
1.0, & \text{if } e = C
\end{cases}
\end{aligned}
\label{eq:discomfort_penalty}
\end{equation}

The discomfort distance $d_{\text{disc}}(O)$ and discomfort penalty factor $p_{\text{disc}}(O)$ are omitted, since the robot does not cause discomfort for static obstacles.

\subsection{Simulator}\label{sec41}

\begin{figure}[!b]
\center{\includegraphics[width=0.74\linewidth]{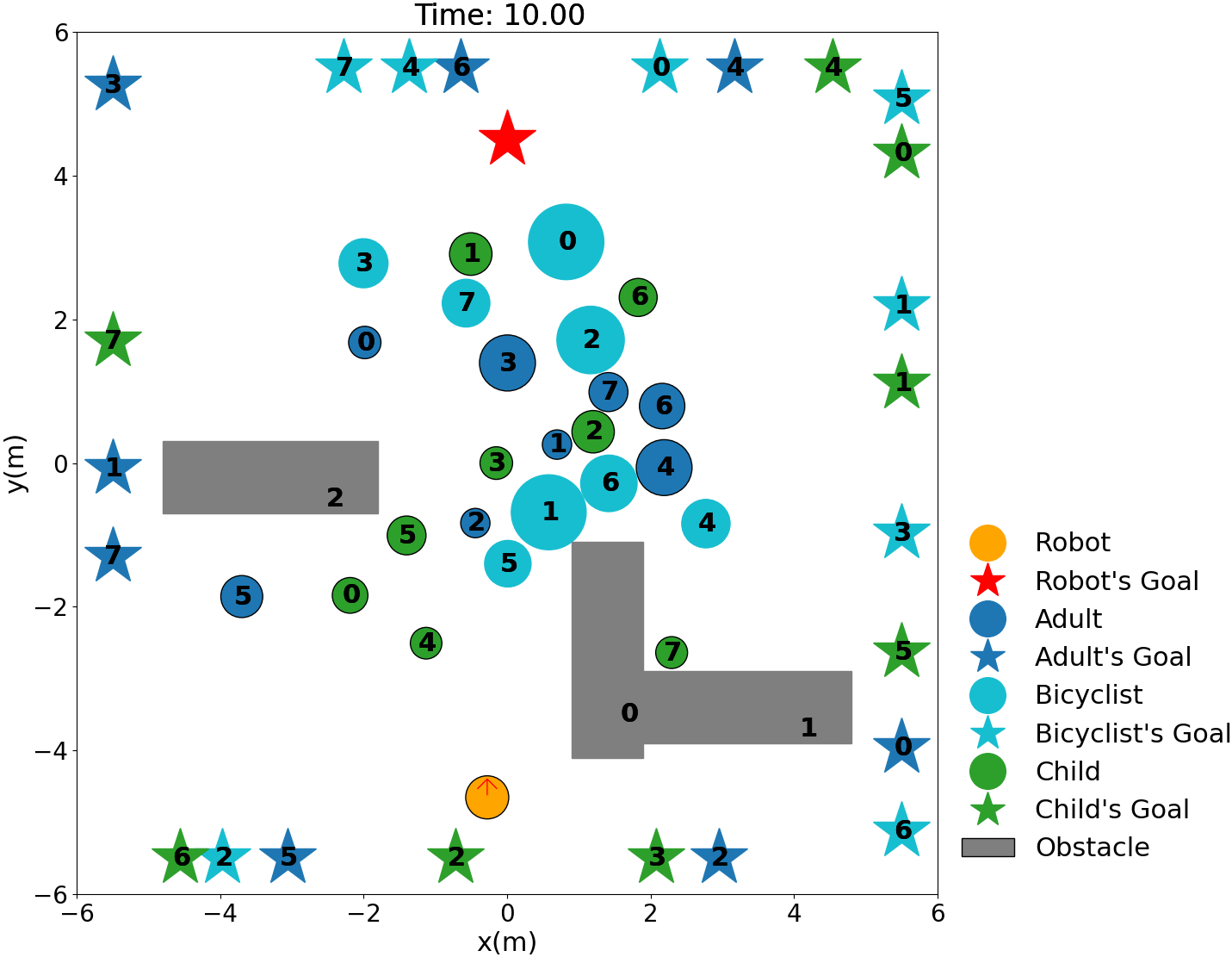}} 
	\caption{A simulator snapshot at $\text{Time} = 10.0$s from the start, showing 8 adults, 8 bicyclists, 8 children, and 3 static obstacles. Static obstacles may overlap.}
	\label{plot_simulator}
\end{figure}

We train and test our models using our simulator. The example of one step of an episode is shown in Figure~\ref{plot_simulator}. The robot, adults, bicyclists, children, and static obstacles are shown. The goals for the robot and dynamic entities are marked as stars. Each entity and the corresponding goals are enumerated.

We use a square crossing scenario, where all the agents are randomly positioned on a square and their goal positions are chosen randomly on the opposite side of the square. Also, we use a circle crossing scenario, where all the agents are randomly positioned on a circle and their goal positions are chosen on the opposite side of the circle.

\subsection{Simulation Setup}\label{sec42}
Simulated agents (adults, children, bicyclists) are controlled by Optimal Reciprocal Collision Avoidance (ORCA) \cite{ORCA}. To introduce diversity, each agent's size and speed are drawn from a uniform distribution specific to its type. The distributions are designed to overlap, for instance, some children may be faster than some adults, making it impossible to determine an agent's type from its size and speed alone. The experiments are conducted in an 11m $\times$ 11m square crossing scenario. Each episode contains one robot, 8 adults, 8 cyclists, 8 children, and 3 static obstacles, all randomly positioned. For simplicity, static obstacles are modeled as stationary pedestrians, with plans to incorporate static grid processing in future work. We compare five navigation methods: CADRL \cite{CADRL}, LSTM-RL \cite{LSTM-RL}, the state-of-the-art SARL \cite{SARL}, a modification called SARL-GP (which adds Time \eqref{eq:time_reward} and Goal Proximity \eqref{eq:proximity_reward} rewards to SARL), and our EB-CADRL. All evaluations use an ``invisible robot'' setting, where other agents do not perceive or react to the robot.
This tests the robot's collision avoidance capabilities in a worst-case scenario that is also realistic, as small robots are often overlooked by people.

\begin{figure*}[!htbp]
	\center{\includegraphics[width=1.0\linewidth]{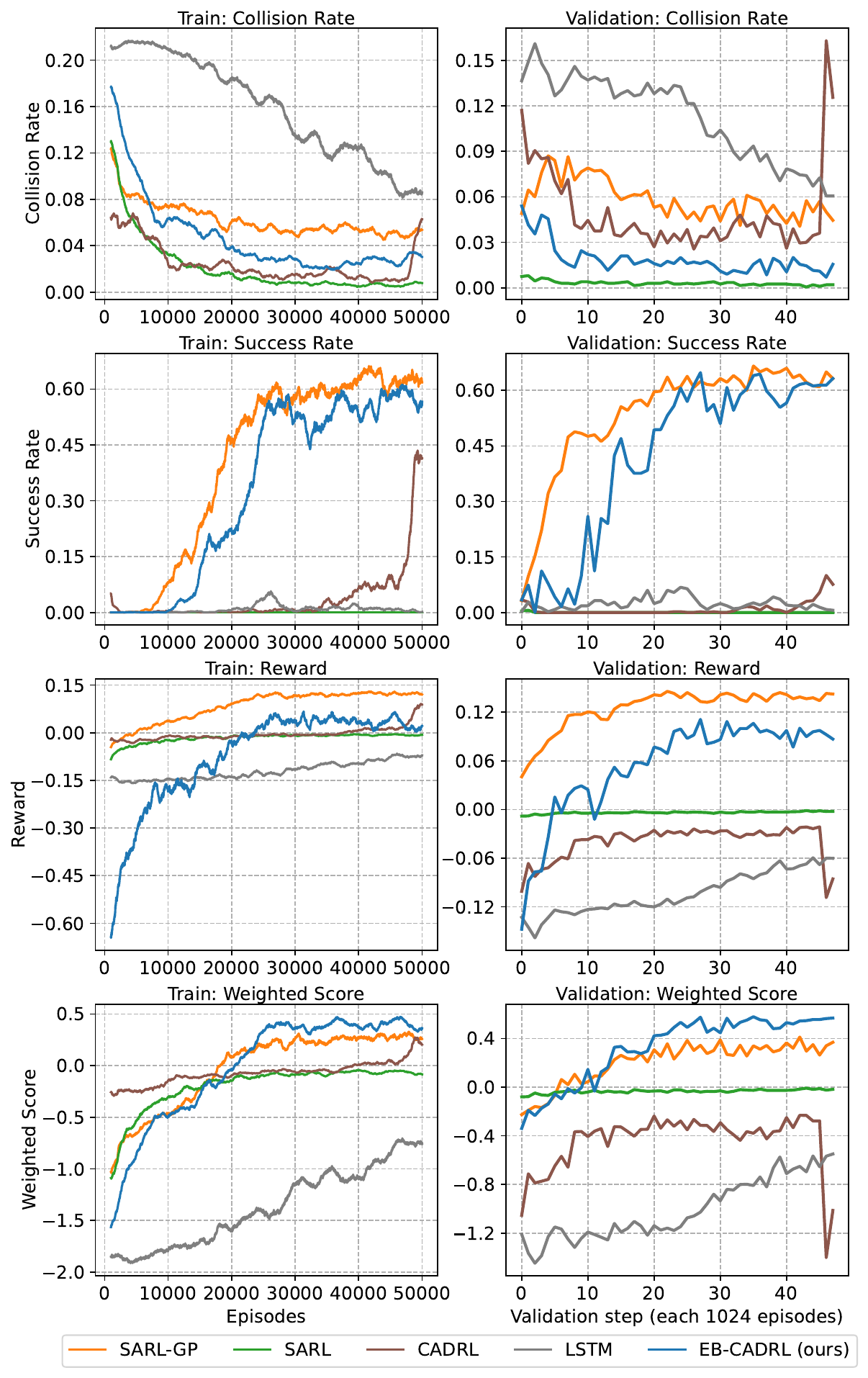}} 
	\caption{Collision rate, success rate, reward, weighted score metrics on the training and validation data. The validation step is performed every 1024 episodes.}
	\label{figure_sr_reward}
\end{figure*}

Models are trained for 3000 episodes with imitation learning, followed by 50000 episodes using the Parallel Deep V-learning algorithm \ref{algorithm_1}. During training, we validate the models every 1024 episodes on a 500-episode validation set. For the final comparison, we select the best-performing model for each method based on its validation performance and evaluate it on a 1000-episode test set.

\subsection{Results}\label{subsec:results}

Figure~\ref{figure_sr_reward} shows the training and validation results. SARL fails to learn a goal-reaching policy, learning only to avoid collisions, and is thus excluded from further analysis.
CADRL and LSTM-RL methods also have poor performance. Additionally, CADRL exhibits an unacceptably high collision rate toward the end of training.

Both SARL-GP and EB-CADRL (ours) successfully converge within 50000 training episodes, reaching a plateau at about 30000 episodes. EB-CADRL outperforms the other methods in both training and validation, as measured by collision rate and weighted score metrics. We compute the Weighted Score $(WS)$ metric using the following formula:
\begin{equation}
\begin{aligned}
WS &= SR - CR(A) - 4.0 \cdot CR(C) - 2.0 \cdot CR(B) - 0.5 \cdot CR(O)
\end{aligned}
\label{eq:weighted_score}
\end{equation}

The coefficients in the Weighted Score metric can be adjusted according to business needs and safety priorities. The current coefficients are chosen to penalize more heavily collisions with more vulnerable entities (children and bicyclists).

The reward metric cannot be used to compare different models as each model has its own reward function. However, it indicates how the method converges and could be used to get the episode for early stopping based on results on the validation dataset.

\begin{table*}[!t]
\centering
\setlength{\tabcolsep}{4pt}
\caption{Metrics on test data. We indicate best metric value in comparison between CADRL, LSTM-RL, SARL-GP and EB-CADRL}
%(since SARL completely fails to learn to navigate to the goal)}
\label{test_table}
% \begin{tabular}{lrrrrr}
\begin{tabular}{lccccc}
\hline
Metric & \multicolumn{1}{c}{SARL} & \multicolumn{1}{c}{CADRL} & \multicolumn{1}{c}{LSTM-RL} & \multicolumn{1}{c}{SARL-GP} & \multicolumn{1}{c}{EB-CADRL} \\
\hline
WS & -0.05 & -0.36 & -0.66 & 0.37 & \bfseries 0.60 \\
SR & 0 & 0.05 & 0.01 & \bfseries 0.68 & \bfseries 0.68 \\
CR & 0.017 & 0.175 & 0.268 & 0.207 & \bfseries 0.068 \\
CR(A) & 0.002 & 0.027 & 0.0485 & 0.027 & \bfseries 0.020 \\
CR(B) & 0.008 & 0.099 & 0.125 & 0.062 & \bfseries 0.018 \\
CR(C) & 0.007 & 0.047 & 0.093 & 0.029 & \bfseries 0.002 \\
CR(O) & 0 & 0.003 & \bfseries 0.002 & 0.089 & 0.028 \\
Time & NaN & 28.62 & \bfseries 27.46 & 29.32 & 29.45 \\
DD(A) & 0.177 & 0.108 & 0.083 & \bfseries 0.130 & 0.118 \\
DD(B) & 0.174 & 0.101 & 0.077 & 0.113 & \bfseries 0.154 \\
DD(C) & 0.170 & 0.113 & 0.083 & 0.131 & \bfseries 0.165 \\
\hline
\hline
\end{tabular}
\end{table*}

In Table \ref{test_table} we provide the result of testing each model on test data which has a size of 1000 episodes. For comparison, we use the following metrics: \textit{SR} - success rate; \textit{CR} - overall collision rate; \textit{CR(A), CR(B), CR(C), CR(O)} - collision rates with adults, bicycles, children and static obstacles, respectively; \textit{Time} - average time to reach the goal; \textit{DD(A), DD(B), DD(C)} - mean danger distances with adults, bicycles and children, respectively. We assume that danger starts when the distance between the robot and the entity is less than 30cm.

SARL has a success rate of 0, indicating that it failed to reach the goal in all attempts.
CADRL and LSTM-RL have very low success rates, making these methods nearly as ineffective as SARL.
SARL-GP and EB-CADRL have a success rate of 0.68, indicating they are similarly effective in reaching the goal.
SARL has the lowest collision rate but failed to reach any goals, indicating it did not navigate successfully enough to encounter many agents and obstacles. EB-CADRL has a significantly lower collision rate compared to CADRL, LSTM-RL and SARL-GP, indicating better safety. Also, EB-CADRL has lower collision rates with adults, bicycles and children compared to CADRL, LSTM-RL and SARL-GP.

\begin{figure*}[!b]
\center{\includegraphics[width=1.0\linewidth]{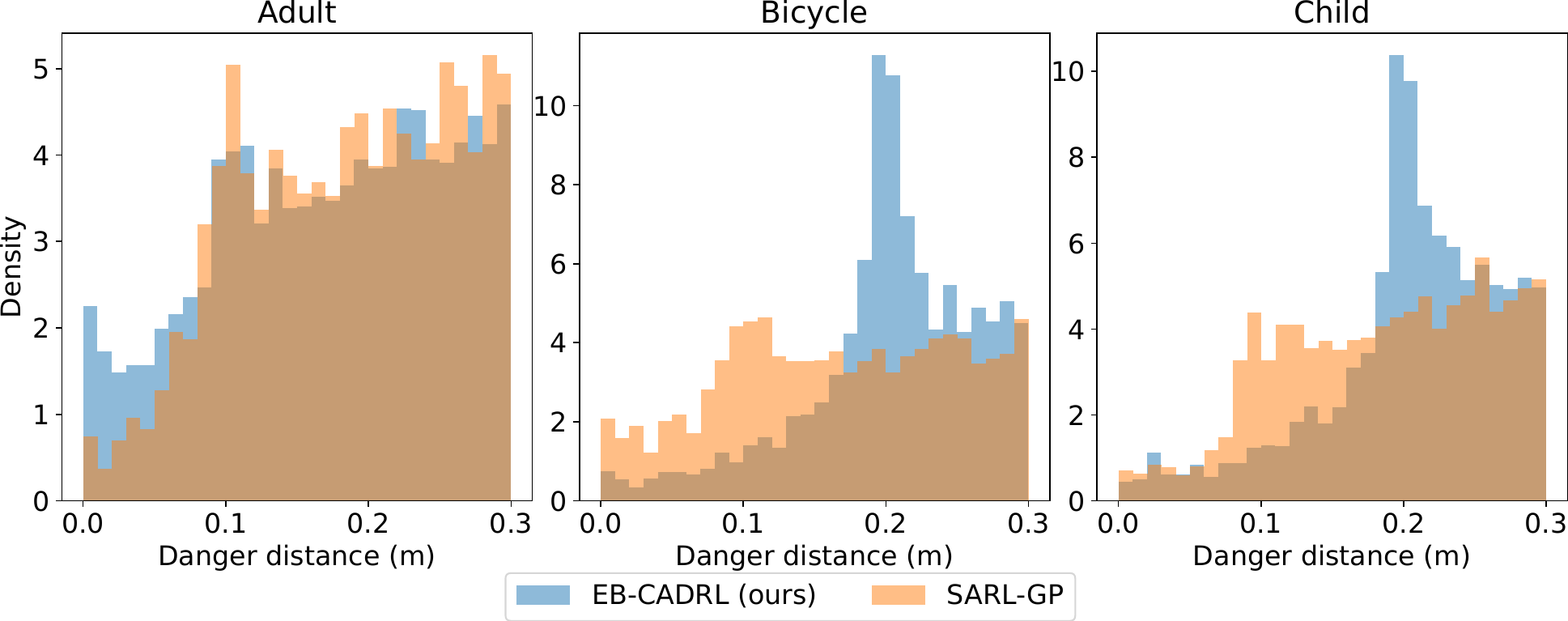}} 
	\caption{Density histograms of danger distances on the test dataset}
	\label{figure_danger_distance_density}
\end{figure*}

SARL-GP and EB-CADRL have similar average times to reach the goal, indicating that EB-CADRL's improved safety does not come at the expense of increased travel time. Time metric for SARL is indicated as NaN, since SARL has a 0 success rate. Although the LSTM-RL shows the shortest average time to reach the goal, this result is less meaningful due to the method's low success rate.

EB-CADRL maintains safer distances from bicycles and children compared to CADRL, LSTM-RL and SARL-GP. Moreover, it significantly outperforms all other methods based on the weighted score metric.

Figure~\ref{figure_danger_distance_density} shows the density histograms of danger distances for SARL-GP and EB-CADRL. EB-CADRL provides enhanced safety when interacting with bicyclists and children, resulting in significantly fewer cases where the robot operates within 0–0.2 meters of these agents compared to SARL-GP.

To sum up, EB-CADRL is the superior method overall because it achieves a good balance between reaching the goal successfully, minimizing collisions, and maintaining safe distances, which makes it efficient and safe for navigation in environments with dynamic and static entities of different types.

\subsection{Ablation Experiments}\label{sec43}

\begin{table}[b]
\centering
\caption{Metrics on test data during ablation experiments}
\label{ablation_table}
\begin{tabular}{lcccccccc}
\hline
 & SR & CR(A) & CR(B) & CR(C) & CR(O) & Reward & Time & Danger time \\
\hline
No entity type & 0.448 & 0.015 & 0.004 & 0.001 & \bfseries 0.002 & 0.037 & 39.078 & 0.913 \\
With entity type & \bfseries 0.501 & 0.016 & \bfseries 0.002 & 0.001 & 0.005 & \bfseries 0.052 & \bfseries 32.857 & \bfseries 0.746 \\
\hline
\end{tabular}
\end{table}

\begin{figure}[!t]
\center{\includegraphics[width=0.7\linewidth]{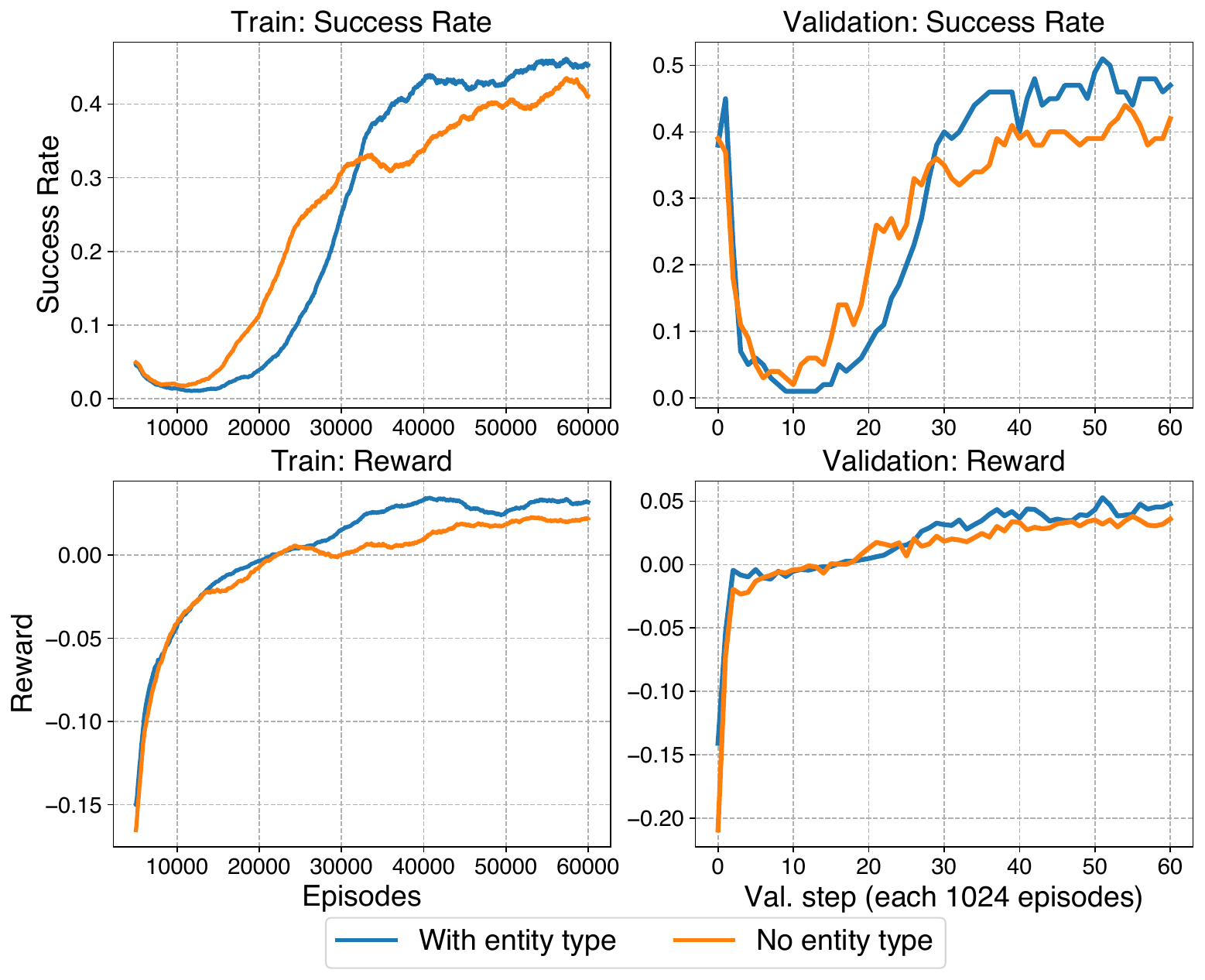}} 
	\caption{Reward and success rate metrics on the training and validation datasets during ablation experiments.}
	\label{figure_ablation}
\end{figure}

To evaluate the impact of including the agent type as a feature, we conducted an ablation study by training and testing two models: one that incorporates the agent type and the other that excludes it. Although the model can infer some aspects of the agent type through features such as radius and velocity, these features alone do not provide a clear distinction between different agent types. For instance, a small, slow-moving cyclist or a large, fast-moving adult could mislead the model. Therefore, we hypothesize that explicitly adding the agent type to the feature set will enhance the model's performance. Both models in this experiment utilize a new reward function and are based on the EB-CADRL.

The results of the ablation experiment on the test dataset are shown in Table \ref{ablation_table}. In addition to the metrics defined in \ref{subsec:results}, we compute the \textit{Danger time} metric: the mean time (in seconds) the robot spent in a potentially dangerous state. A lower value indicates safer navigation and less discomfort caused to other entities.

The model incorporating the entity type outperforms the one excluding it in several key aspects. It achieved a higher reward and better success rate, indicating superior overall performance. Additionally, it registered fewer bicycle collisions, reached the goal faster, and spent less time in danger. However, the model excluding the entity type had slightly fewer collisions with adults and obstacles.
The reward and success rate metrics on the training and validation data are shown in Figure~\ref{figure_ablation}.

\section{Conclusion}\label{sec:conclusions}

In this study, we introduced Entity-Based Collision Avoidance using Deep Reinforcement Learning (EB-CADRL), a novel approach for autonomous robot navigation in environments with dynamic agents and static obstacles. To enhance safety and efficiency, EB-CADRL leverages an entity-specific reward function that ensures more context-aware navigation by penalizing the robot differently based on entity type. Our experimental results demonstrate that EB-CADRL significantly outperforms state-of-the-art techniques, achieving a high success rate with a lower collision rate across various entity types. The model also effectively balanced safety and efficiency, maintaining safe distances from entities without significantly increasing travel time. Furthermore, ablation experiments validated the critical role of incorporating entity-specific information, which consistently improved performance across reward, success, and safety metrics. This result underscores the value of our context-aware approach in developing more sophisticated autonomous navigation systems.

Future work could be devoted to enhancing the model's robustness in environments with occlusions and complex agent behaviors, integrating additional entity types into the model, such as cars and motorcycles, exploring multi-robot coordination, and conducting real-world testing.

\newpage

\bibliographystyle{splncs04}
\bibliography{bibliography.bib}

\begin{thebibliography}{10}
\providecommand{\url}[1]{\texttt{#1}}
\providecommand{\urlprefix}{URL }
\providecommand{\doi}[1]{https://doi.org/#1}

\bibitem{AutonomousRobots}
Bekey, G.A.: Autonomous Robots: From Biological Inspiration to Implementation and Control. The MIT Press (2005)

\bibitem{ORCA}
van~den Berg, J., Guy, S.J., Lin, M., Manocha, D.: Reciprocal n-body collision avoidance. In: Pradalier, C., Siegwart, R., Hirzinger, G. (eds.) Robotics Research. pp. 3--19. Springer Berlin Heidelberg, Berlin, Heidelberg (2011)

\bibitem{CHARALAMPOUS201785}
Charalampous, K., Kostavelis, I., Gasteratos, A.: Recent trends in social aware robot navigation: A survey. Robotics and Autonomous Systems  \textbf{93},  85--104 (2017)

\bibitem{SARL}
Chen, C., Liu, Y., Kreiss, S., Alahi, A.: Crowd-robot interaction: Crowd-aware robot navigation with attention-based deep reinforcement learning. In: 2019 International Conference on Robotics and Automation (ICRA). pp. 6015--6022 (2019)

\bibitem{socially_aware_2017}
Chen, Y.F., Everett, M., Liu, M., How, J.P.: Socially aware motion planning with deep reinforcement learning. In: 2017 IEEE/RSJ International Conference on Intelligent Robots and Systems (IROS). pp. 1343--1350 (2017)

\bibitem{CADRL}
Chen, Y.F., Liu, M., Everett, M., How, J.P.: Decentralized non-communicating multiagent collision avoidance with deep reinforcement learning. In: 2017 IEEE International Conference on Robotics and Automation (ICRA). p. 285–292 (2017)

\bibitem{drew2021multi}
Drew, D.S.: Multi-agent systems for search and rescue applications. Current Robotics Reports  \textbf{2},  189--200 (2021). \doi{10.1007/s43154-021-00048-3}

\bibitem{LSTM-RL}
Everett, M., Chen, Y.F., How, J.P.: Motion planning among dynamic, decision-making agents with deep reinforcement learning. In: 2018 IEEE/RSJ International Conference on Intelligent Robots and Systems (IROS). pp. 3052--3059 (2018)

\bibitem{SocialAwareRobot_2013}
Ferrer, G., Garrell, A., Sanfeliu, A.: Social-aware robot navigation in urban environments. In: 2013 European Conference on Mobile Robots. pp. 331--336 (2013)

\bibitem{GEHRKE2023100789}
Gehrke, S.R., Phair, C.D., Russo, B.J., Smaglik, E.J.: Observed sidewalk autonomous delivery robot interactions with pedestrians and bicyclists. Transportation Research Interdisciplinary Perspectives  \textbf{18},  100789 (2023)

\bibitem{social_force}
Helbing, D., Moln\'ar, P.: Social force model for pedestrian dynamics. Phys. Rev. E  \textbf{51},  4282--4286 (May 1995). \doi{10.1103/PhysRevE.51.4282}

\bibitem{Kretzschmar_2016}
Kretzschmar, H., Spies, M., Sprunk, C., Burgard, W.: Socially compliant mobile robot navigation via inverse reinforcement learning. The International Journal of Robotics Research  \textbf{35} (01 2016). \doi{10.1177/0278364915619772}

\bibitem{KRUSE20131726}
Kruse, T., Pandey, A.K., Alami, R., Kirsch, A.: Human-aware robot navigation: A survey. Robotics and Autonomous Systems  \textbf{61}(12),  1726--1743 (2013)

\bibitem{lavalle_2006}
Lavalle, S.M.: Planning Algorithms. Cambridge University Press (2006)

\bibitem{mnih2015humanlevel}
Mnih, V., et~al.: Human-level control through deep reinforcement learning. Nature  \textbf{518}(7540),  529--533 (Feb 2015). \doi{10.1038/nature14236}

\bibitem{Pytorch}
Paszke, A., et~al.: Pytorch: an imperative style, high-performance deep learning library. In: Proceedings of the 33rd International Conference on Neural Information Processing Systems. Curran Associates Inc., Red Hook, NY, USA (2019)

\bibitem{SpringerHandbook}
Siciliano, B., Khatib, O.: Springer Handbook of Robotics. Springer Handbooks, Springer (2016). \doi{10.1007/978-3-319-32552-1}

\bibitem{Sutton}
Sutton, R.S., Barto, A.G.: Reinforcement Learning: An Introduction. A Bradford Book, Cambridge, MA, USA (2018)

\bibitem{navigation_raw_depth_inputs_2018}
Tai, L., Zhang, J., Liu, M., Burgard, W.: Socially compliant navigation through raw depth inputs with generative adversarial imitation learning (05 2018)

\bibitem{UnfreezingtheRobot_2010}
Trautman, P., Krause, A.: Unfreezing the robot: Navigation in dense, interacting crowds. pp. 797--803 (09 2010). \doi{10.1109/IROS.2010.5654369}

\bibitem{Xue2024}
Xue, B., et~al.: Crowd-aware socially compliant robot navigation via deep reinforcement learning. International Journal of Social Robotics  \textbf{16}(1),  197--209 (2024)

\end{thebibliography}

\end{document}